\documentclass{article}
\usepackage{arxiv}

\usepackage[T1]{fontenc}
\usepackage{graphicx}
\usepackage{amsmath}   % \text{} in math mode (provided by the jair class before)
\usepackage{amssymb}

\makeatletter
\let\ax@title\title
\renewcommand{\title}[2][]{%
  \ax@title{#2}%
  \def\ax@tmp{#1}%
  \ifx\ax@tmp\@empty\else\renewcommand{\shorttitle}{#1}\fi
}
\gdef\@author{}
\newcommand{\ax@add}[1]{\g@addto@macro\@author{#1}}
\newif\ifax@firstauthor \ax@firstauthortrue
\renewcommand{\author}[1]{%
  \ifax@firstauthor\ax@firstauthorfalse\else\ax@add{\And}\fi
  \ax@add{#1}%
}
\newcommand{\authornote}[1]{\ax@add{\thanks{#1}}}
\newcommand{\authornotemark}[1][]{}
\newcommand{\orcid}[1]{}
\newcommand{\email}[1]{\ax@add{\\ \texttt{#1}}}
\newcommand{\affiliation}[1]{\ax@add{\\ #1}}
\newcommand{\institution}[1]{#1}
\newcommand{\city}[1]{\\ #1}
\newcommand{\state}[1]{, #1}
\newcommand{\country}[1]{, #1}
\newcommand{\shortauthors}{}
\newcommand{\setcopyright}[1]{}
\newcommand{\acmDOI}[1]{}
\newcommand{\JAIRAE}[1]{}
\newcommand{\JAIRTrack}[1]{}
\makeatother
\newcommand{\eg}[1]{(\textit{e.g.} #1)}

\setcopyright{cc}
\acmDOI{}

\usepackage{hyperref}

\RequirePackage[
  style=authoryear,
  natbib=true,
  backend=biber,
  giveninits=true,
  uniquename=init
  ]{biblatex}

\usepackage[capitalize]{cleveref}
\usepackage{array}
\usepackage{multirow}
\begin{document}

%%
%% The ``title'' command has an optional parameter,
\title[Evidence of Concept Mastery in LLMs]{Evidence of Concept Mastery in Large Language Models: Generalized v. Memorized Rule Application}
% \title[Satzilla]{SATzilla: Portfolio-based Algorithm Selection for SAT}

%% used to denote shared contribution to the research and/or corresponding author.
\author{José Luiz Nunes}
% \authornote{Corresponding Author.}
\orcid{0000-0002-8215-6150}
\email{jnunes@inf.puc-rio.br}
\affiliation{%
  \institution{Pontifical Catholic University of Rio de Janeiro}
  \city{Rio de Janeiro}
  \state{Rio de Janeiro}
  \country{Brazil}
}

\author{Guilherme de Franca C F Almeida}
\authornote{Corresponding author.}
\orcid{0000-0002-9134-9843}
\email{guilhermefcfa@insper.edu.br}
\affiliation{%
  \institution{INSPER}
  \city{São Paulo}
  \state{São Paulo}
  \country{Brazil}}

\author{Brian Flanagan}
\orcid{0000-0001-7719-5051}
\email{brian.flanagan@mu.ie}
\affiliation{%
  \institution{Maynooth University}
  \city{Maynooth}
  % \state{British Columbia}
  \country{Ireland}
}

% The short list of authors must be made of the list of all authors' lastnames.
\renewcommand{\shortauthors}{Nunes, Almeida \& Flanagan}
\renewcommand{\shortauthors}{Anon}

%% In the article class \maketitle must precede the abstract.
\maketitle

\begin{abstract}
\textbf{Background.}
Previous research has put forward evidence that large language models (LLMs) reproduce human judgments. However, this does not by itself establish conceptual mastery: the correspondence may reflect memorisation or sensitivity to incidental task features.

\noindent \textbf{Objective.}
Across five experiments, we test whether LLMs possess a generalisable competence in applying rules, including in hard cases in which a rule's text and purpose point towards different outcomes.

\noindent \textbf{Method.}
We present the results of five rule violation experiments spanning 13 different LLMs.
Studies~1A and~1B compared LLM judgments with newly collected human data ($n = 115$), using both published stimuli and matched vignettes created after the models' training cut-offs.
Studies~2A and~2B used pre-existing human data to test responses to time-pressure instructions, a manipulation with a mechanistic route to human judgment that is blocked for LLMs.
Study~3 varied reasoning effort in models that expose it as a parameter, in an attempt to use an equivalent mechanism to constrained time human judgements.
Studies~1B and~2B additionally varied system prompt wording and numerical scale anchors.
Model temperature was calibrated per model against the response variance of the human sample.

\noindent \textbf{Results.}
LLM judgments closely tracked human judgments for both published stimuli and matched vignettes created after the training cut-offs of the models tested in Study~1A.
Humans and LLMs also responded in the same unanticipated direction to differences between the two sets, placing less weight on rule text in the novel cases.
Sensitivity to text and purpose remained robust across changes in system prompts and numerical scale anchors.
Responses to time-pressure instructions were mixed and model-specific: some models ignored cues describing constraints that altered the resources available to humans but not to LLMs, suggesting a distinction between conceptual competence and human alignment.
Replication of the human pattern was most apparent in models with fewer parameters, and these effects were substantially more susceptible to changes in system prompt and scale anchors than the effects of text and purpose.
Increasing reasoning effort produced no detectable change in rule application for most reasoning models, and sensitivity to text and purpose remained stable, though a significant purpositivist trend was observed in higher-effort settings for GPT-oss and Claude Sonnet~5.
Response variance remained lower for LLMs than for humans despite our per-model temperature calibration.

\noindent \textbf{Conclusions.}
Overall, the findings are inconsistent with rote memorisation or prompt-specific responding alone and suggest that LLM rule application reflects a generalisable, standing semantic competence that does not typically depend on expanded deliberation.

\textbf{Keywords:}
Large Language Models; Rule-based decision-making; Conceptual mastery; Experimental Jurisprudence; Legal reasoning.
\end{abstract}

\section{Introduction}

%\subsection{Concept Mastery}
Large Language Models (LLMs) have matched (or exceeded) human performance across a wide range of tasks, from cybersecurity \cite{aisi-gpt, aisi-mythos}, to fiction writing \cite{DoshiHauser2024GenerativeAIEnhances}, and from medical exams \cite{nori_capabilities_2023} to general conversation \cite{jones_people_2024}. Critics, however, object that what looks like mastery in each of these domains might be the result of simple memorisation \cite{magar_data_2022, wang_generalization_2024, ma-etal-2025-memorization-understanding}.

One challenge in teasing apart mastery and memorisation is that LLM cognition, like that of humans, is opaque. First, that means that both humans and LLMs are unreliable introspectors. Asking someone to introspect on the functioning of their own visual system is not likely to lead to scientific insight into how light hitting the human eye interacts with complex neural circuitry to produce visual experience. Similarly, asking an LLM for an explanation of their abilities is bound to mislead. Second, for both the human brain and for LLMs, we lack a complete explanation of how the lower-level components generate their complex behaviours (see \citeauthor{sharkey2025openproblemsmechanisticinterpretability}). Thus, even when fine-grained data about the LLMs underlying architecture is available, we might have trouble disentangling rote memorization from genuine generalisation.

These challenges are well known to cognitive scientists, who developed several tools to investigate human behaviour that rely upon neither first person report nor complete mechanistic explanations. Such methods are increasingly being used to yield insight into the operation of generative AI. Recent work in machine psychology \cite{hagendorff_machine_2023} has investigated LLMs' morality \cite{dillion_can_2023, park_diminished_2023, ji_moralbench_2024, nunes_are_2024}, their ability to render causal judgments \cite{nie_moca_2023}, their theory of mind \cite{strachan_testing_2024, GandhiEtAl2023UnderstandingSocialReasoninga} as well as several other concepts \cite{almeida_exploring_2024}. In each case, researchers have found that LLMs can produce results that are highly similar to those generated by humans. Importantly, LLMs have been shown to reproduce not only instances of rational human behaviour, but also biases \cite{malberg-etal-2025-comprehensive, almeida_exploring_2024, acerbipnas}. Finally, although very similar on aggregate, LLMs' responses were still found to be significantly different from those produced by humans due to an effect that has been dubbed ``diminished diversity of thought'' \cite{park_diminished_2023}: repeated sampling of a single LLM leads to significantly less variance than repeated sampling from a human population. This has led to overall exaggerated effect sizes for machine psychology experiments performed on LLMs \cite{almeida_exploring_2024}.

One methodological issue with previous work comparing humans and LLMs on rule violation judgment is that diversity of thought in LLMs is partially determined by a tunable parameter: temperature. While different models are affected by temperature in different ways, lower levels will result in the model prioritizing the token it has learned as most likely to follow a given text. On the other hand, higher values increase the chance of using tokens assigned lower probability, a behaviour usually described as creative or exploratory. Previous research either set LLM temperature to zero in order to ensure reproducibility \cite{guha_legalbench_2023, BlairStanekVanDurme2026LLMsProvideUnstable}, left it at default \cite{sachdeva_normative_2025}, or chose arbitrary points for different models \cite{almeida_exploring_2024, nunes_are_2024, franken_procedural_2024, shen_mind_2025}. All of these strategies make it difficult to assess whether the systematic variations observed between the diversity of human responses and those of different LLMs are due to (a)~architectural differences or (b)~the specific temperatures used.\footnote{A different alternative, consisting of using the log probability of the N most likely tokens \cite{Choi2024}, is either unavailable \eg{for Gemini}, or returns unreliable output (see discussion in \href{https://community.openai.com/t/logprobs-inconsistent-between-runs-for-4o/935082}{https://community.openai.com/t/logprobs-inconsistent-between-runs-for-4o/935082}) in current SOTA models.}

Producing matched diversity of thought would provide more clarity on the magnitude and direction of potentially existing differences between LLM- and human-generated responses. Nonetheless, the issue of memorisation would persist: the exact wording of stimuli employed in research with humans is often available on the open web and is therefore likely to feature in the pre-training data for major LLMs. Therefore, instead of actually mastering the relevant concepts, LLMs may simply be reproducing memorized patterns included in their training datasets, which could explain not only the striking similarities between human- and LLM-generated responses, but also why LLMs show diminished variance. Further bolstering the sceptic's case, LLMs have been shown to be sensitive to features that would be irrelevant for those who have mastered the relevant concepts and abilities, such as subtle changes in prompt wording and deployment context \cite{guha_embroid_2023, guha_legalbench_2023, rottger_political_2024, beck_sensitivity_2023, SclarEtAl2024QuantifyingLanguageModels, gupta-etal-2024-self, schroder2025largelanguagemodelssimulate, ErricaEtAl2025WhatDidWrong}. For instance, previous work has reported a tendency for LLMs to choose the same numerical value even in reversed scales, where meaning has completely inverted \cite{park_diminished_2023}.

Given this background, a central question in current machine psychology is whether LLMs have genuinely acquired conceptual competences that were once exclusive to human cognition, or function merely as heavyweight autocompletes. Competence presupposes that LLM responses are both robust to irrelevant differences and sensitive to relevant ones. More broadly, these studies illustrate a behavioural strategy for evaluating candidate LLM competences. Evidence should converge across generalisation to novel cases, sensitivity to relevant semantic variation, robustness to incidental task features, and stability across changes in inference conditions. Human alignment is neither necessary nor sufficient. Where a manipulation changes human cognitive resources but not those available to a model, competent performance may involve principled divergence rather than mimicry.

Legal reasoning provides the test domain. LLMs are increasingly deployed in legal practice, making this application especially salient. Moreover, the burgeoning literature on the experimental jurisprudence of rules offers a suite of different stimuli and research designs that can be adapted for machine psychology. Across three studies, comprising five experiments and spanning 13 LLMs, we find that models remain sensitive to both rule text and purpose across differently worded stimuli, system prompts, and response scales. Moreover, we find that time-pressure instructions produced mixed, model-specific effects rather than reliably reproducing the human pattern, further weakening a rote-memorisation account. Finally, we find that LLMs typically apply rules without depending on expanded inferential reasoning, consistent with fluent deployment through standing semantic competence.

\subsection{Rules and Legal Reasoning in humans and LLMs}
\label{subsec:rules}
The task of rule application plays a significant role in social interaction, most prominently in legal contexts \cite{schauer_thinking_2012}. Recent work in jurisprudence has employed experiments to uncover the concept of rule. This strategy has revealed that people rely on both a rule's letter and spirit when making violation judgements, as well as on several further factors which selectively amplify their respective effects (see \citeauthor{almeida_rules_2025} for an overview). For instance, one vignette tells participants about a rule prohibiting shoes in an apartment that was introduced to keep the floor clean. Participants are then presented with cases where the rule's text and purpose either a)~both imply that the rule has been violated \eg{someone walks in with dirty boots}, b)~both imply that the rule has not been violated \eg{ someone walks in barefoot with clean feet}, c)~diverge, with text implying that the rule has been violated and purpose implying the opposite \eg{someone walks in with brand new shoes}, or d)~diverge, with purpose implying that the rule has been violated, but text implying the opposite \eg{someone walks in barefoot but with very dirty feet}.

Previous work replicated some of these studies with LLMs, revealing very high correlations between rule applications by humans and LLMs \cite{almeida_exploring_2024}. These results align with related research indicating that LLMs perform comparably to humans in identifying utterances' pragmatic interpretation \cite{HuEtAl2023FinegrainedComparisonPragmatic} and in recognizing a difference between conduct that violates a rule outright and conduct that exploits a loophole \cite{MurthyEtAl2023ComparingEvaluationProduction}. However, the stimuli used had been available on the open web as of the LLMs' training data cut-off. This raises the aforementioned worry that the LLMs might have simply reproduced patterns represented in their training set for those same examples, instead of truly deploying the relevant concept. Moreover, they also display the aforementioned ``diminished diversity of thought'' typical of LLMs \cite{park_diminished_2023} where variance across different answers of the same model is less prominent than that observed in a comparable human sample.

\begin{figure*}[t]
\begin{center}
\includegraphics[width=\textwidth, trim = 0cm 3.5cm 0cm 3.5cm, clip]{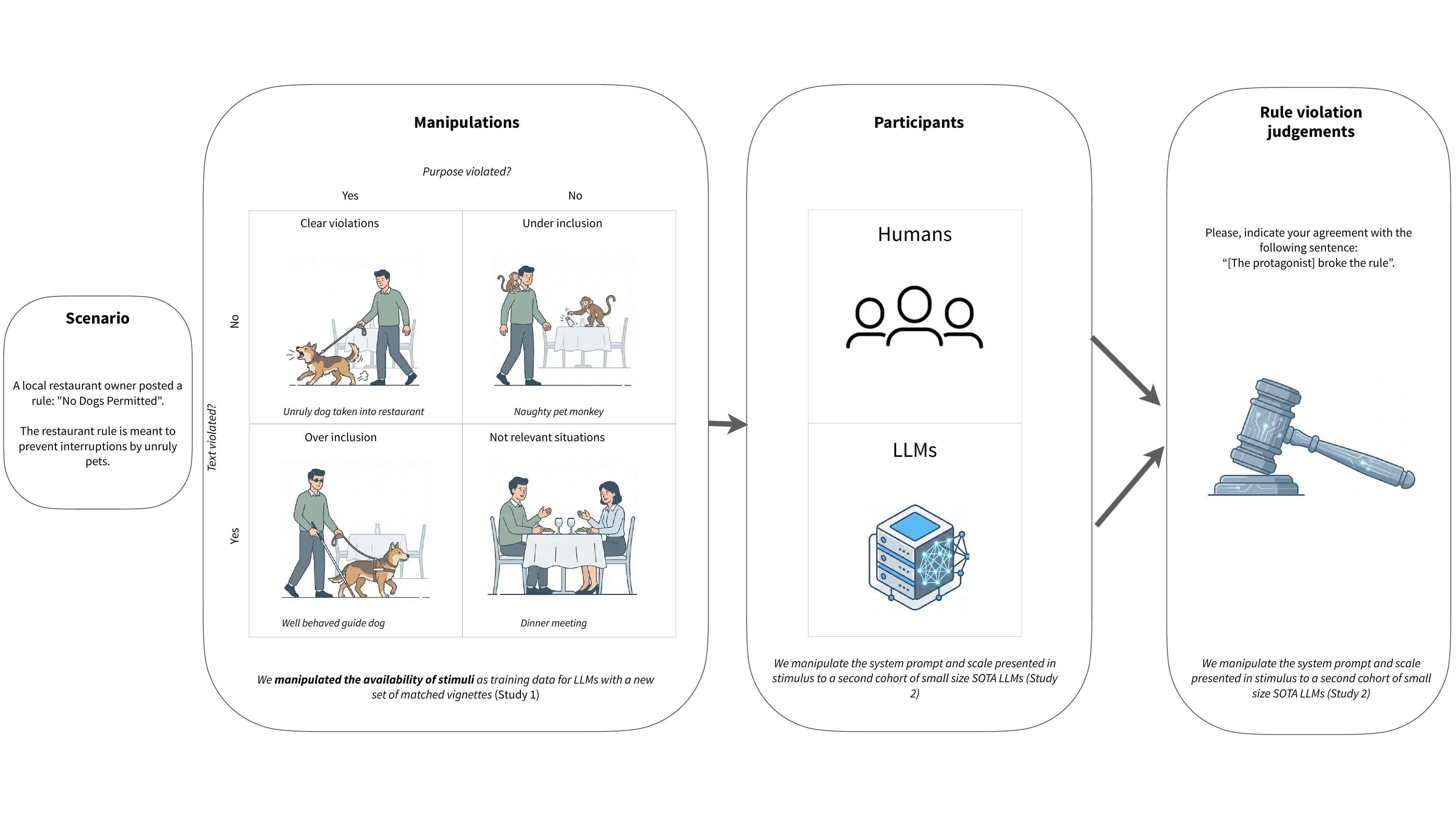}
\end{center}
\caption{We conducted five experiments on rule violation manipulating: availability of stimuli as training data (Study~1a and 1b), time pressure for responding (Study~2a and 2b), the wording of model instructions (Study~1b and 2b), and reasoning effort level (Study 3). We found LLMs' judgements tracked human ones even where the latter varied unexpectedly in response to a new set of vignettes, and were robust to our prompt manipulations. This figure illustrates stimuli for experiments 1a and 1b, but images were not used as stimuli.}
\label{fig:overview}
\end{figure*}

\section{Study~1A}

\begin{table}[htbp]
\centering
\caption{Original and New Vignettes}
\label{tab:vignettes}

\begin{tabular}{
>{\centering\arraybackslash}m{0.14\linewidth}
>{\centering\arraybackslash}m{0.14\linewidth}
p{0.34\linewidth}
p{0.34\linewidth}
}
\hline

{\small\textbf{Text violated}} 
& {\small\textbf{Purpose violated}} 
& {\small\centering\arraybackslash\textbf{Original vignettes}} 
& {\small\centering\arraybackslash\textbf{New vignettes}} \\

\hline

-
& -
& A local restaurant owner posted a rule: ``No Dogs Permitted''.

The restaurant rule is meant to prevent interruptions by unruly pets.
& Restaurant management posted a notice: ``Dogs are forbidden''.

The rule was introduced to stop noisy pets disturbing customers. \\

\hline

Yes & Yes
& Joe enters the restaurant at lunchtime with his unruly pet dog.
& Vicky enters the restaurant accompanied by her noisy pet dog. \\

\hline

Yes & No
& Derek, who is blind, enters the restaurant with his well-trained guide dog.
& Louis, who has a disability, walks into the restaurant with a well behaved service dog. \\

\hline

No & Yes
& Bill enters the restaurant bringing with him his naughty pet monkey.
& Paula enters the restaurant together with her squawking pet parrot. \\

\hline

No & No
& Steve and some colleagues have a meeting over dinner at the restaurant.
& Josh takes his friend Lara to have lunch at the restaurant. \\

\hline

\multicolumn{4}{>{\centering\arraybackslash}p{0.96\linewidth}}{
Please, indicate your agreement with the following sentence:

``[The protagonist] broke the rule''.
} \\

\hline
\end{tabular}
\end{table}

The most direct and convincing rote memorization objection points to the fact that all of the stimuli for the LLM studies mentioned in \cref{subsec:rules} were available online prior to the models' training cut-off date (See \cref{tab:models-overview}). Suppose that a teacher were to give their students an exam with the exact same questions that were made available as homework. In that case, a student's good performance on the exam could be equally well explained by memorisation and by learning. In contrast, if the exam contained different questions that probed the same underlying knowledge, a student's performance on it would be a much more reliable indicator of learning. Similarly, if LLMs respond to a set of stimuli it has seen before in the same way that humans do, this could be explained by either memorisation or generalisation. But if they match humans on unseen stimuli, that would be much better evidence of the latter.

Study 1A tests this hypothesis by replicating a study on rule violation judgments \cite{FlanaganEtAl2023MoralAppraisalsGuide} with a novel set of stimuli that was not available online until after the tested LLMs training cut-off date. We propose that, if LLMs and humans respond to novel stimuli in the same way that they do to the original materials, these results cannot be explained by rote memorization alone.

\subsection{Methods}
\label{sec:methods-1a}

We recruited 120 participants from Prolific. After excluding 2 incomplete responses and 3 participants who failed our \href{https://aspredicted.org/ck95-whg7.pdf}{pre-registered attention check}, we were left with a final human sample of 115 respondents (36 male; 79 female; mean of age $= 36$). All data, stimuli, and analysis scripts are available at: \url{https://osf.io/uvy9x/?view_only=95db33761f92421393ed9d58c46131cb}

We also generated responses to the same stimuli using 240 separate instances of each of the following LLMs: Llama~3.2 90b, Gemini Pro, Claude~3 Opus, and GPT-4o. For GPT-4o, after discarding answers that failed the pre-registered attention check replicated from the original study or were not proper answers within the scale, we were left with 2784 responses compared to 3840 from other models\footnote{We dropped 992 rows from GPT-4o in which the model had failed the attention check question. We then also removed 64 rows where we could not detect a valid answer to the experiment within the defined scales.}. Human data collection and generation finished on January of 2025, while the latest training cut-off for the aforementioned model was May 2024.

Replicating previous research \cite{FlanaganEtAl2023MoralAppraisalsGuide}, Study~1A followed a 2 (text: violated, not violated) $\times$ 2 (purpose: violated, not violated) $\times$ 4 (scenario: no dogs in the restaurant, no vehicles in the park, no shoes inside, no shooting wild animals) within-subjects design, with a between-subjects manipulation of novelty: participants were presented with either the original, old vignettes (which were already published on the open web) or with matched vignettes, which we created for this specific study. Each participant/model instance received all 16 versions of either the new or old vignettes. See \cref{tab:vignettes} for sample vignettes.

\subsection{Temperature Calibration}
To determine the temperature setting least likely to artificially produce diminished diversity of thought for each LLM, we computed the human sample's standard deviation for each unique cell in the ``original vignettes'' condition. We then discretized each model temperature range into 10 values, starting from 0, and generated 10 responses to the relevant stimuli with each model and each of the 10 temperatures. For each of those responses, we created a vector with the standard deviation for each of the 16 unique cells in the experimental design and calculated the mean squared error between it and that of the human sample.

For the LLMs deployed in Study~1A, the temperatures which achieved the minimum error vis-a-vis human standard deviation were 1.0 for Llama 3.2 90b, 0.9 for Gemini Pro, 0.4 for Claude 3, and 1.8 for GPT-4o.

%For the second set of LLMs, deployed in Study ?, the temperatures which achieved the minimum error vis-a-vis human standard deviation were...  As temperature variation was not supported (or not recommended) for Gemini~3 Flash and GPT-5 Mini, for these models we instead varied the applicable ``reasoning effort''. The values that most resembled human diversity-of-thought for each model were: $reasoning\_effort=none$ for Gemini~3 and $reasoning\_effort=minimal$ for GPT~5.

\subsection{Results}

\begin{figure}[ht]
\begin{center}
\includegraphics[width=.7\columnwidth]{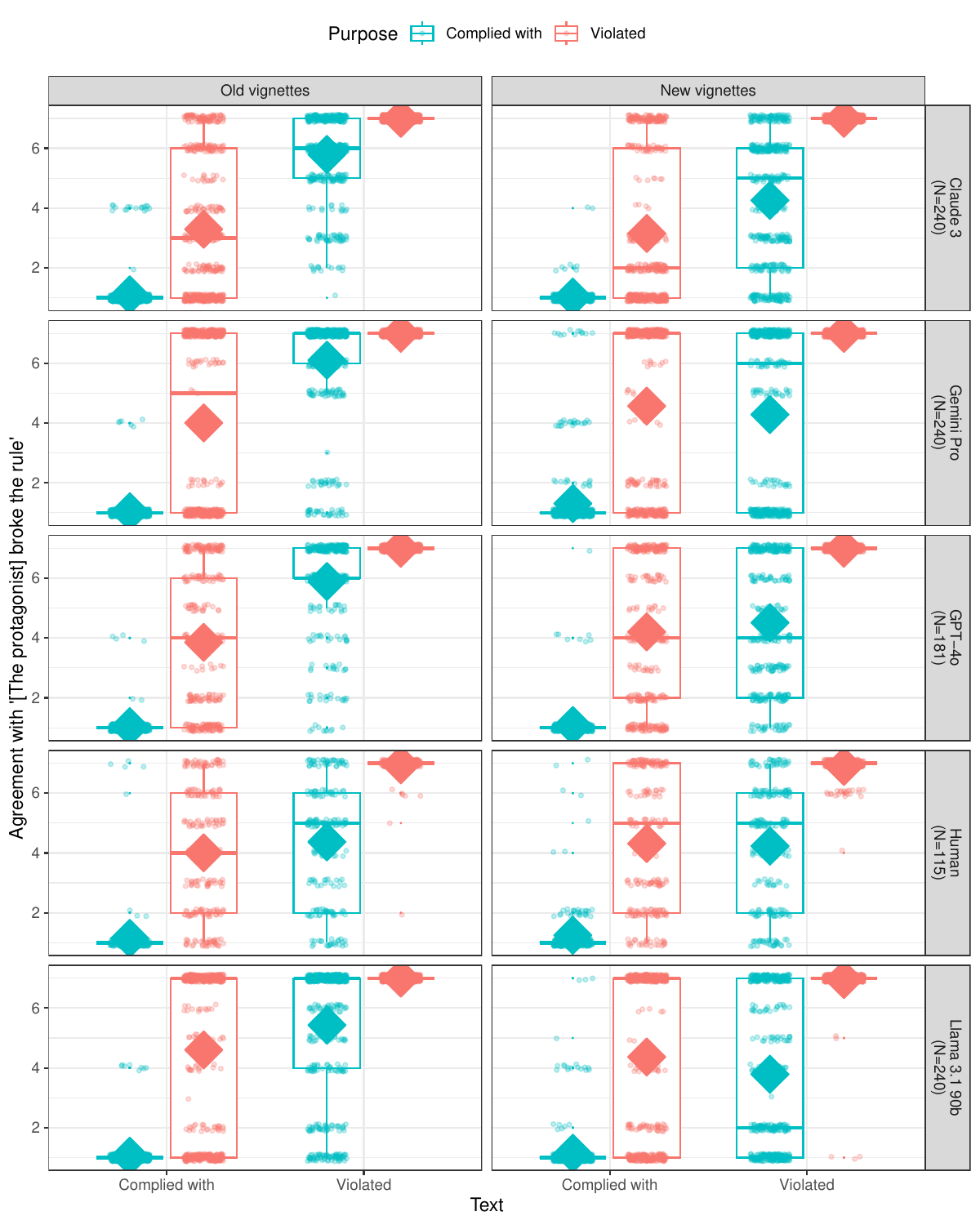}
\end{center}
\caption{Distribution of rule violation judgement per agent.}
\label{fig:dist_study1}
\end{figure}

In line with our \href{https://aspredicted.org/ck95-whg7.pdf}{pre-registered analysis plan}, we fit individual mixed effects models with fixed effects for text, purpose, novelty, and all interactions, while accounting for random effects of participant/model instance. As predicted, these mixed effects models revealed that both text and purpose were significant predictors of rule violation judgements for all models (all $p\text{s} < .001$), as depicted in \cref{fig:dist_study1}. Also in accordance with our predictions, there were no significant main effects of novelty -- that is, whether participants were evaluating new or old vignettes -- over rule violation judgements among humans ($F_{(1, 1716)} = 0.57$, $p = 0.453$). This finding was qualified by an unexpected significant interaction between text and novelty ($F_{(1, 1716)} = 4.4$, $p = .036$) reflecting text's reduced influence on participant responses to new vignettes ($B=2.78$, $t=27.58$, $p_{\text{Tukey}} < .001$) compared to old vignettes ($B=3.08$, $t=29.79$, $p_{\text{Tukey}} < .001$)\footnote{We acknowledge this is a small difference. Normalizing with the residuals SD, it represents a difference of $0.2$, and a semi-partial $R^2=0.0007$ using the r2glmm library. However, we highlight this difference as its effects were significant and large with LLMs as reported ahead.}.

Contrary to our expectations, main effects of novelty were significant for all LLMs; as were all unanticipated interactions involving novelty (all $p\text{s} < .001$). Surprisingly, simple slopes analysis revealed that the text*novelty interaction was consistently in the same direction as that observed among humans, with all LLMs displaying significantly (all $p\text{s}_{\text{Tukey}} < .001$) larger effects of text for the old ($4.2 > \text{all } B\text{s} > 3.08$) as compared to the new vignettes ($3.54 > \text{all } B\text{s} > 2.62$), with an opposite trend for the purpose*novelty interaction, with significantly (all $p\text{s}_{\text{Tukey}} < .001$) more purposivism observed for new vignettes ($3.19 > \text{all } B\text{s} > 2.43$) when compared to the old ($2.56 > \text{all } B\text{s} > 1.7$).

As set out in the pre-registration, we also ran an ANOVA based on a mixed effects model with fixed effects of text, purpose, novelty, and agent, as well as all interactions, while accounting for random effects of scenario and participant. The analysis confirmed the predicted existence of significant differences between models, with significant main effects of agent ($\chi^2_{(4)} = 96.39$) qualified by the predicted two-way interactions with text ($\chi^2_{(4)} = 182.575$) and purpose ($\chi^2_{(4)} = 141.4$, all $p\text{s} < .001$). All other terms in the model also achieved statistical significance ($p < .001$).

Finally, an exploratory analysis of the per-cell standard deviations of human and LLM-generated data, controlling for random effects of scenario, showed a significant effect of agent on variation of responses. Pairwise comparisons revealed Claude~3 ($\overline{s}=0.59$, $p_{\text{Tukey}} < .02$) and GPT-4o ($\overline{s}=0.573$, $p_{\text{Tukey}} < .012$) showed significantly less variance than human responses ($\overline{s}=1.16$), while the comparisons between humans and Gemini~Pro ($\overline{s}=0.695$) and Llama~3 ($\overline{s}=0.7$) were not significant. 

\subsection{Discussion}

The results of Study~1A are inconsistent with the hypothesis that similarity in human and LLM rule violation judgements is caused by memorization of published human responses to particular stimuli. The same factors that affect human rule violation judgements (text and purpose) were found to affect LLMs even when they applied rules to scenarios created after the data cut-off date for model training. Moreover, in a striking and unanticipated result, just as the new vignettes elicited lesser adherence to text among humans than the originals, so too did they elicit lesser adherence among LLMs. This suggests that LLMs can generalise from their training data to pick out even subtle differences that are hard to articulate.

Regarding response variance, Claude~3 and GPT-4o still displayed significantly less variation even after our calibration. Equally, some models in our sample gave invariant responses in reaction to some of our stimuli (see online Supplementary Materials). These results suggest that, while our calibration had partial success in increasing variability in the responses of models, some aspects of diminished variety of thought are resistant to temperature tuning.

While indicative of concept mastery, the results of Study~1A are not sufficient to allay all concerns critics have expressed. As mentioned in the introduction, previous research has documented the influence of the wording of the system prompt on response patterns \cite{schroder2025largelanguagemodelssimulate, gupta-etal-2024-self, rottger_political_2024, beck_sensitivity_2023, petersen2026sense}. In addition, the order in which scale items are presented has been shown to influence LLM judgments \cite{gupta-etal-2024-self, park_diminished_2023}. Thus, it could be that the high correlations between human intuitions and LLM-derived judgements observed in this study were caused by the specific prompting strategy we employed.

\section{Study~1B}

For this study, we made three important changes in the process of generating data based on the materials of Study 1A. First, as some of the models used in Study~1A have been deprecated, we updated to a cohort of more recent LLMs, opting to survey smaller and cheaper models which are more likely to be deployed at scale in legal decision-making. Second, we manipulated the system prompt: responses were generated using either a) one of two short system prompts with no instructions as to how to behave (``You are a participant in a study about rule violation.'' and ``You are a helpful assistant that acts as a participant in psychological studies.''),
\footnote{The absence of instructions constraining the response format resulted in multiple formats for responses. We parsed these results, while excluding responses that contained more than one valid rating.
% Non-systematically, we observed instances where generated text initially provided one answer, which was later changed it along reasoning traces.
}
or b) one of two similar, longer alternatives, including the original system prompt used in Study~1A, which adds instructions such as ``Your response SHOULD NOT contain the number's accompanying text. So, if you select `7', you should just return `7', instead of `7 - Strongly agree'.''

Our goal was to test the robustness of the results observed in the previous study when varying instructions, the order of answer options \cite{gupta-etal-2024-self, schroder2025largelanguagemodelssimulate}, and a new set of models. As the new set of vignettes was available online before data collection for this experiment, we cannot assert the novelty is in Study~1A.

Finally, we manipulated the scale labels, employing both the original option, ranging from ``1 - Strongly disagree'' to ``7 - Strongly agree'', and a reversed version ranging from ``1 - Strongly agree'' to ``7 - Strongly disagree''. If the results observed above are created by selective attachment to certain numeric anchors in reaction to some types of vignettes \eg{selecting 6 or 7 for cases in which text was violated}, this manipulation should have a high impact on our results.

\subsection{Methods}
\label{sec:methods-1b}

Following the same procedure of Study~1A described in \cref{sec:methods-1a}, we queried each LLM 240 times, 120 with the original set of vignettes from \cite{FlanaganEtAl2023MoralAppraisalsGuide} and 120 from the ones crafted for Study~1A. We generated data from the following models: Qwen~3 (\texttt{qwen3-235b-a22b-2507}), Claude 4.5 Haiku(\texttt{claude-haiku-4-5-20251001}), Gpt-5 Mini (\texttt{gpt-5-mini-2025-08-07}), Gemini~3 Flash (\texttt{gemini-3-flash-preview}).

\subsubsection{Temperature Calibration}

The temperatures which achieved the minimum error vis-a-vis human standard deviation were:$0.9$ for Qwen~3 and $0.8$ for Claude~4.5 Haiku. As temperature variation was not recommended (or not supported) for Gemini~3 Flash and GPT-5 Mini, for these models we instead varied the applicable ``reasoning effort''. The values that most resembled human diversity-of-thought for each model were: \texttt{reasoning\_effort=none} for Gemini~3 and \texttt{reasoning\_effort=minimal} for GPT~5.

\subsubsection{Data processing}

We again discarded answers where the output was not a valid answer within the scale, or if it contained two different values from our scale since that would produce ambiguous results without qualitative analysis. We also interrupted generation of scenarios when the model did not provide a valid answer for the attention check -- thus none of the 4 scenarios were rated. The total number of scenario ratings our condition could produce was 30560 for each model. As a result of our validation, we were left with 30033 rows for Claude~4.5 Haiku, 25797 for GPT-5 Mini, 18714 for Gemini~3 Flash, and 30292 for Qwen~3\footnote{We interrupted the generation for runs in which the attention check did not contain the number 2. Then, we dropped 527 rows for Haiku, 295 for Gemini, 126 for Qwen 3 and 4750 for GPT-5 Mini due to not identifying a valid answer for either the vignette or the attention check. Finally, we dropped 13 additional rows from GPT 5 mini and 142 from Qwen~3 that failed the attention check.}.

\subsection{Results}

\begin{figure}[t]
\begin{center}
\includegraphics[width=0.7\columnwidth]{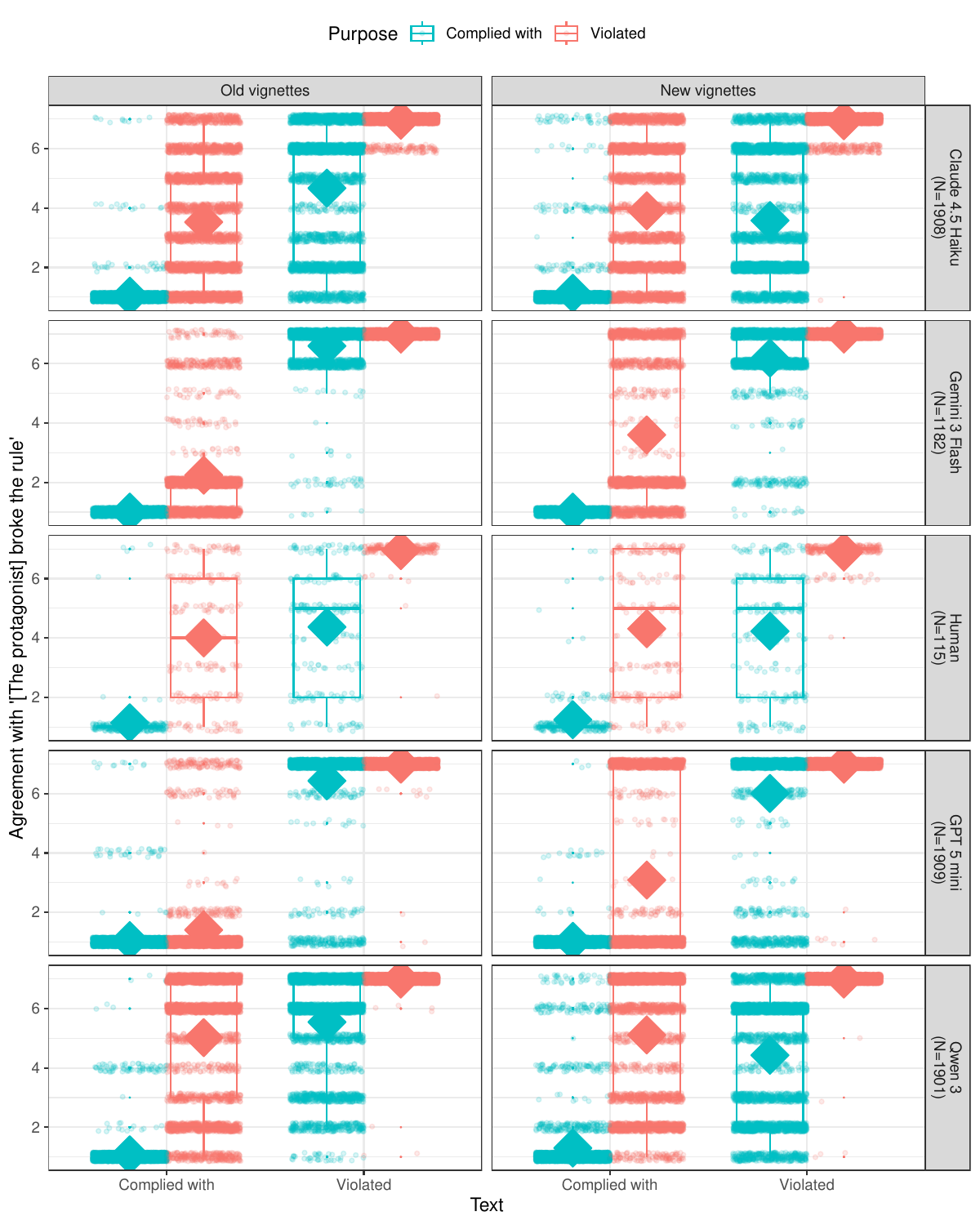}
\end{center}
\caption{Distribution of rule violation judgements for second set of models across all prompt and scale manipulations.}
\label{fig:dist_exp2}
\end{figure}

To check whether the effects found on Study 1A replicated for all variations of system prompt and scale order, we modelled rule violation judgments as a function of text, purpose, and the text*purpose interaction, allowing each of those effects to interact with predictors for novelty, LLM (designating which model generated that response), system prompt, and scale formulation, while accounting for random effects of LLM instance and scenario. The specified ANOVA revealed the expected significant effects of text ($X^2_{(1)} = 197070$, $p < .001$) and purpose ($X^2_{(1)} = 61946.21$, $p < .001$), but these were qualified by significant interactions with each other, as well as two- and three-way interactions involving novelty, system prompt, scale formulation, and LLM. The only non-significant interaction was the three way interaction of text $\times$ purpose $\times$ scale ($X^2_{(1)} = 1.034$, $p = .309$).

In the above-mentioned model, the estimated marginal effects of text and purpose were significant ($p_s < .0001$) and in the expected direction ($ 0.8 < \text{all } \beta_{s}\text{purpose} < 1.19$, $ 1.75 <$ all $\beta_s\text{text} < 2.72$) for all levels of novelty, system prompt, and scale formulation. To ensure the effects were also robust for each different LLM, we ran multiple ANOVAs of text, purpose, and the text:purpose interaction while accounting for random effects of model instance and scenario for each LLM. Each ANOVA was fit to a subset of the data determined by a single value of either novelty, system prompt, or scale. This meant we fitted the specified model eight times per LLM using different subsets. For all fitted models, the simple effects of text and purpose were significant (all $p\text{s} < .001$) and in the predicted direction. The text:purpose interaction was significant in all but 3 subsets, all from Claude Haiku~4.5.

Turning to an investigation of the unanticipated effects of novelty over text and purpose, we sought to replicate Study 1A's finding that, for both humans and LLMs, text became less important in the new vignettes in comparison to the original set. For all LLMs, Tukey-corrected pairwise comparison inspecting the interaction terms revealed that text became less important and purpose more important in the new vignettes when compared to the original set (all $|z|\text{s} > 14.37$, all $p\text{s} < .001$).

\begin{figure}[ht]
\begin{center}
\includegraphics[width=0.8\columnwidth]{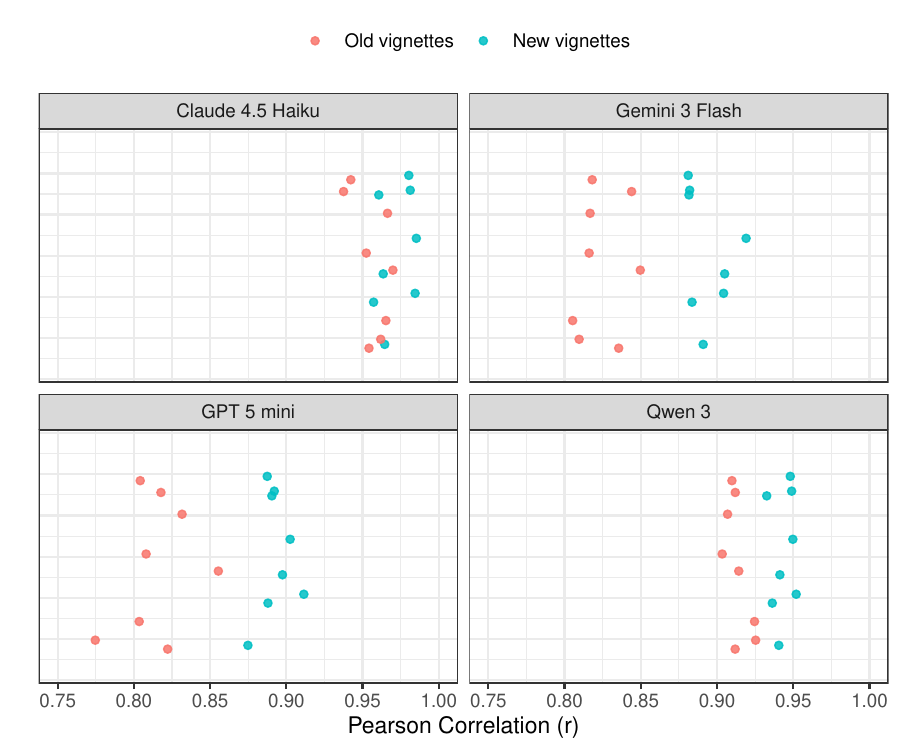}
\end{center}
\caption{Distribution of correlations for each model by condition in Study 1b.}
\label{fig:exp2_corr}
\end{figure}

Next, we deployed a similar strategy to test whether the correlation between LLM-derived judgments and human intuition varied substantially with these manipulations. First, we averaged judgments over each combination of scenario, text, purpose, and novelty, resulting in 32 unique cells per LLM and for the human sample.
For each condition, we then computed the correlation between the cells generated by each LLM with the ones derived from human participants for each combination of system prompt and scale order resulting in 16 correlations per model.
All correlations were significant ($\text{all } p_{s}\text{Holm-Bonferroni}<.002$) and can be seen in \cref{fig:exp2_corr}. There was little variation in the overall correlation of responses, with the lowest coefficient observed in GPT5 ($r=0.775$), with the median at $r=0.9085$.

As seen in \cref{fig:exp2_corr}, the correlations were overall higher on the new vignettes. To test whether that difference between old and new vignettes was statistically significant, we ran a set of exploratory t-tests comparing the average correlation between conditions for each model. All pairwise comparisons were significant ($p_s\text{Holm-Bonferroni} < .02$) and in the same direction, with Claude displaying the lowest difference.

Finally, we carry the same exploratory analysis of the standard deviation in responses, calculated for each cell across scenario $\times$ purpose violation $\times$ text violation $\times$ novelty. However, the result of this analysis changes if we collapse all variations of system prompt and scale, or maintain them as distinct observations. In the first case, pairwise comparisons were significant ($p_s{\text{Tukey}} < .0001$) only for Gemini ($\overline{s}=0.55$) and Qwen ($\overline{s}=0.59$). On the other hand, all comparisons are significant when we consider prompt and scales as different cells (all pairwise $|\beta| > 0.39$, $p_s{\text{Tukey}} < .03$). In all cases, variance was lower for LLMs than for humans.

\subsection{Discussion}

Study 1B explored the robustness of rule violation judgments rendered by LLMs to putatively irrelevant variation in system prompt wording and scale anchors. While both of these factors did significantly affect LLM rule application (which was reflected in significant interactions involving those terms), the effects were not sufficient to crowd out the markers of conceptual competence: just as in Study 1A, text and purpose were still reliably associated with rule violation. Thus, Study 1A's findings do not seem to be an artefact of a specific system message or a preference for certain points within a Likert scale. LLMs' capability to track human intuition in rule violation judgments proved robust to our manipulations, suggesting that machine access to the meaning of a rule is not confined to the rule's letter, but extends also to its spirit. 

The results also extend Study 1A's findings - including the unanticipated preference for text on the original set of vignettes - to a more recent and smaller class of LLMs, which includes one open weight model (Qwen). This extension shows that even cheaper and faster models that are more likely to be deployed at scale in legal settings are capable of reliably tracking human judgment in rule application.

\section{Study~2A}

Studies~1A and 1B give us reason to believe LLMs can generalize concepts beyond their training set, sorting their application according to scenarios' semantic content. Another way to test the memorisation vs. generalisation question pertains to how models react to features of the stimuli that should elicit different responses from humans, but not from LLMs. Among humans, rule violation judgments have been shown to be sensitive to time pressure: participants forced to delay their response for 15 seconds display an increased tendency to rely on text as the criterion for rule violation judgments when compared to participants forced to respond within 4 seconds \cite{FlanaganEtAl2023MoralAppraisalsGuide}. We can make sense of this effect, as the cognitive process for deciding instantly differs from that required to answer a question after a forced delay. In contrast, our procedure for testing LLMs attempts to ensure (with a high degree of success) that the models output exactly one token in response to each question. This means that, unlike humans, the LLMs we are testing process our input for the exact same time regardless of whether that input includes instructions that communicate time pressure or a forced delay. Thus, a time pressure manipulation will have no effect on the actual time that the AI takes to accomplish the task. In other words, the path through which this manipulation exerts its influence on human judgments is ``blocked'' for LLMs.

If all that the LLM is doing is reproducing content it has memorised, we should expect that the vignettes employed by \citet{FlanaganEtAl2023MoralAppraisalsGuide} should elicit higher reliance on purpose when paired with instructions communicating time pressure and induce text-based responses when paired with instructions mentioning a forced delay. After all, that is the pattern that was available for the LLMs during their pre-training. On the other hand, if models are generalizing beyond that simple memorisation step, they should recognize that the instructions are irrelevant.

To make the study suitable for LLMs, we adapted the stimuli and procedure from \citet{FlanaganEtAl2023MoralAppraisalsGuide}, which followed a 2 (text: violated, not violated) $\times$ 2 (purpose: violated, not violated) $\times$ 2 (condition: speeded, delayed -- between-subjects) design. To acclimate human participants to the time pressure and forced delay conditions, the original research featured a training round in which information was displayed either for a limited time only, or for a fixed time before the participant could proceed. This taught participants the experiment's interface. Since this was unnecessary for LLMs, we skipped the training round.

Following the original procedure, we announced at the beginning of the experiment that participants would have to respond either in a ``short time'' (speeded condition) or would be ``given time to think'' (delayed condition). We could neither display questions to LLMs for a short time only, nor impose on them an extended period of deliberation. Instead, after each vignette, we added a reminder informing them of the time available to humans in the relevant condition. In the speeded condition we included ``Remember: you will only have 4 seconds to answer''; in the delayed condition we added the following text ``Remember: you must reflect 15 seconds before answering''.

In the original experiment, participants assigned to the delayed condition were asked to justify their answers in a short paragraph. We chose to omit this part of the stimulus, since the reflection instruction could elicit chain-of-thought prompting, which has been shown to significantly impact LLM performance \cite{kojima_large_2022, wei_chain--thought_2022}. Instead, we were interested in checking whether LLM rule application could be influenced by a manipulation that did not affect the number of tokens generated: across both conditions, valid responses included only a single token -- the numerical indicator of rule violation within the scale.

\subsection{Methods}
We generated 240 runs for each model also used in Study~1A (Llama~3.2 90b, Gemini Pro 1.5, Claude~3 Opus, and GPT-4o. See \cref{sec:methods-1a}), choosing randomly between the 36 different groups from Study~1 from \citet{FlanaganEtAl2023MoralAppraisalsGuide}.
\footnote{We highlight we made one deviation from the original procedure to generate data for Studies~2A and 2B. The original study used a latin-square design with 12 groups for the speeded conditions, each participant rating 8 scenarios, and 24 for the delayed condition, each participant rating 4 scenarios. Due to a mistake in our code, we swapped the group allocation and used the 12 groups with 8 scenarios for delayed condition with LLMs, and the 24 with 4 for speeded. This choice had been made to balance time for participants under the two conditions in the original study. Thus, although this leaves us with an unplanned difference from the original study, we do not believe this to materially change the results.}
As in the previous study, we included an attention check in our protocol, and discarded any generations that provided the wrong answer.

After an initial analysis of the amount of valid results, we found the GPT-4o often failed to produce a valid answer to the attention check, resulting in only 154 initial valid answers. Thus, we decided to run another 120 instances of our protocol with this model. This resulted in 228 valid answers for GPT-4o, compared to 240 from Claude, Gemini, and Llama 3.2.

\subsection{Results}

\begin{figure}[ht]
\begin{center}
% Placeholder for Figure 2
\includegraphics[width=.9\columnwidth]{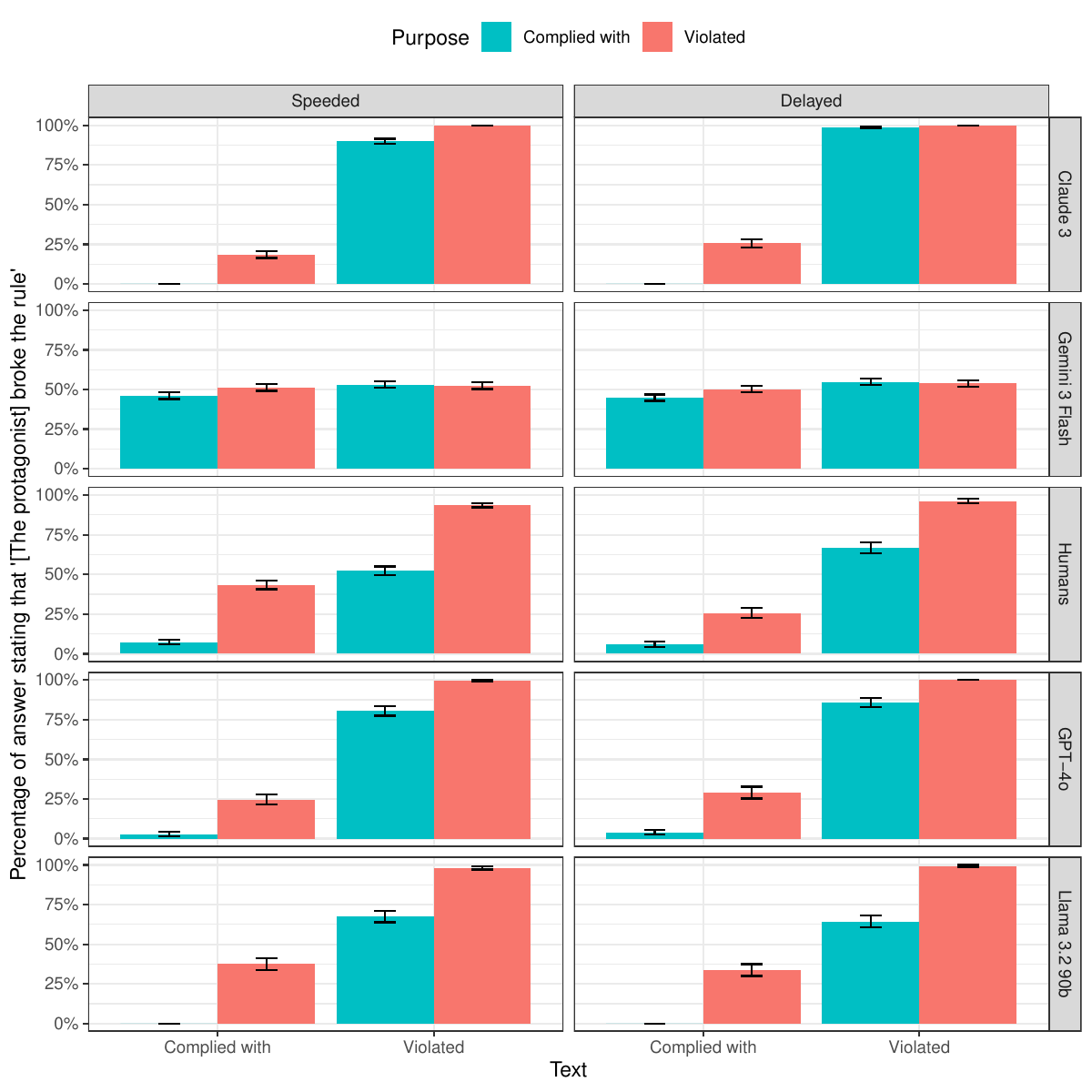}
\end{center}
\caption{Percentage of answers that agreed that the participant broke the rule by condition.}
\label{fig:initial_speeded_delayed_comparison}
\end{figure}

To analyse the data for Study~2A, we followed the strategy laid out in \citet{FlanaganEtAl2023MoralAppraisalsGuide}, fitting mixed effects models of rule violation judgment with fixed effects for text, purpose, condition, and all interactions between them while accounting for random effects of participant (model instance) and scenario and then ran ANOVAs based on this specification for each LLM.

As reported in \citet{FlanaganEtAl2023MoralAppraisalsGuide}, rule violation judgments among humans were affected by text ($F_{(1, 1790)} = 1082.73, p < .001$), purpose ($F_{(1, 1789)} = 419.09, p < .001$), the text*purpose interaction ($F_{(1, 1790)} = 4.37, p = .037$), as well as by significant interactions between text and condition ($F_{(1, 1791)} = 27.72, p < .001$) and purpose and condition ($F_{(1, 1790)} = 17.61, p < .001$). The latter two interactions were such that the effects of text were stronger in the delayed condition ($B = 0.657, t = 23.75, p < .001$) when compared to the speeded condition ($B = 0.476, t = 23.36, p < .001$), while purpose exerted more influence in the speeded condition ($B = 0.386, t = 18.97, p < .001$) than in the delayed condition ($B = 0.242, t = 8.76, p < .001$).

For GPT-4o and Llama 3.2 90b, although we did find significant main effects of both text (GPT: $F_{(1, 957)} = 2201.03, p < .001$, Llama: $F_{(1, 1056)} = 1213.70, p <.001$) and purpose (GPT: $F_{(1, 959)} = 165.18, p < .001$, Llama: $F_{(1,1055)} = 346.59, p <.001$), no other effects reached statistical significance (all $F\text{s} < 2.5$, all $p\text{s} > 0.11$), indicating that these models' rule violation judgments were not influenced by the time pressure manipulation.

In contrast, the individual models fitted to the data produced by Gemini Pro~1.5 and Claude 3 did reveal significant main effects of condition (Gemini: $F_{(1, 202)} = 7.16, p = .008$, Claude: $F_{(1, 199)} = 8.62, p = .0037$) qualified by significant three-way interactions with text and purpose (Gemini: $F_{(1, 1040)} = 5.57, p = .0185$, Claude: $F_{(1, 2248)} = 18.67, p < .001$). Inspecting marginal effects, we could see that this was driven by a significant difference in the way the models treated one class of cases: those where only text was violated, as can be observed in the differences between the second blue bar in each column of \cref{fig:initial_speeded_delayed_comparison}.

These LLMs -- just as the human participants -- were significantly more likely to classify cases where only text was violated as breaking the rule in the delayed when compared to the speeded condition (Gemini: $M_{\text{delayed}} = 0.70, M_{\text{speeded}} = 0.54, B = 0.16, t = 4.42, p_{\text{Tukey}} < .001$; Claude: $M_{\text{delayed}} = 0.98, M_{\text{speeded}} = 0.89, B = 0.08, t = 4.39, p_{\text{Tukey}} < .001$). However, unlike humans, Gemini treated cases where only purpose was violated the same in the two conditions ($M_{\text{delayed}} = 0.49 , M_{\text{speeded}} = 0.46, B = 0.03, t = 0.78, p_{\text{Tukey}} = 0.99$). For Claude, whereas there was a significant difference between speeded and delayed cases where only purpose was violated, this difference ran in the opposite direction to that observed among humans, with the LLM lending less weight to purpose under speeded ($M_{\text{speeded}} = .18$) than under delayed conditions ($M_{\text{delayed}} = 0.24, B = 0.06, t=3.25, p_{\text{Tukey}} = .0262$).

As with Study~1A, we also fitted a single mixed effects model of rule violation judgments across all agents with text, purpose, condition, LLM, and all two- three- and four-way interactions between them as fixed effects while accounting for random effects of scenario and participant. An ANOVA based on this specification revealed that all terms exerted statistically significant influence over rule violation judgements (see Supplementary Materials for full results). Importantly, this includes main effects for LLM, as well as all interactions involving LLM, which shows that response patterns varied significantly between humans and LLMs, as well as between different LLMs.

Once again, all LLMs outputted responses with significantly lower variance than humans ($X^2_{(4)} = 73.26, p < .001$), despite using the temperature settings derived for Study 1A.

\subsection{Discussion}

GPT-4o and Llama 3.2 90b were unaffected by our time pressure manipulation, lending the same weight to text and purpose whether instructed to respond quickly or after a pause for deliberation. In contrast Gemini Pro and Claude 3 had their responses change in reaction to the instructions. Moreover, this happened in different ways for different models. By increasing the value assigned to text in the delayed as compared to the speeded condition, both models exhibited the same response as humans with regards to text-only violations. But they did not behave the same with regards to cases which violated only purpose. Whereas humans are more likely to think this class of case violates the rule when applying it under time pressure, Claude is more likely to say the opposite.

Presumably, time pressure causes humans to respond differently because it deprives them of the cognitive resources necessary for them to competently apply their concept. Tasked with determining the most likely next token, an LLM must determine whether to produce the token that a human in such conditions would be expected to express or instead to produce the token that an entity immune to such conditions (like itself) would be expected to express. In that case, by paying equal attention to text in the speeded condition, thereby acting as if they are not deprived of resources on which a correct answer might depend, GPT and Llama exhibit concept mastery. Accordingly, these models might be thought to correctly identify the concept as it appears to agents with sufficient information processing capacity. On the other hand, Gemini Pro and Claude 3 showed an unexpected sensibility to the time pressure manipulation. That might be partially explained by memorization, as the LLMs acted similarly to humans in becoming less likely to say that text-only violations broke the rule under speeded condition. But even that would be an incomplete explanation, as Claude's decreased sensitivity to purpose-only violations under time pressure contrasts with the human pattern. Thus, overall, the results of Study 2A speak against the hypothesis of memorisation.

\section{Study~2B}
Study 2B tests whether the time-pressure effects observed in Study 2A survive when the newer model cohort is subjected to the prompt- and scale-robustness manipulations used in Study 1B. For this task, the scale used was a binary choice. As the form of scale modification, we swapped the number that represented whether the protagonist broke the rule from the original scale (``1 - Yes, 2 - No'') to (``2 - Yes, 1 - No'') without changing the order presented. Again, if models are capable of behaviour compatible with understanding this variation should not affect results.

\subsection{Methods}

We queried each one of Claude 4.5 Haiku, Qwen~3, Gemini~3 Flash, and GPT-5 Mini (the same cohort of models and parameters described in \cref{sec:methods-1b}) 120 times for the delayed condition, and 120 for the speeded condition under each combination of $scale \times system\_prompt$.

\subsubsection{Data Processing}

As in the previous study, we discard answers that were ambiguous, outside the scale, as well as responses where the attention check was failed. Since we attribute group randomly, we do not have a set number of expected answers to be generated. We again interrupted runs that did not contain the correct digit for the valid attention check. After processing, we had 10183 answers for Qwen~3, 9963 for GPT-5 Mini, 10319 for Gemini 3 Flash, and 10096 for Claude 4.5 Haiku\footnote{We started with 10220 rows for Qwen~3, 10137 for GPT-5 Mini, 10320 for Gemini 3 Flash, and 10157 for Claude 4.5 Haiku. After validating the extraction for each one we have the respective number of valid rows: 10183, 10023, 10319, and 10096. Upon removing rows that failed the attention check, we arrive at the number reported in the main text.}

\subsection{Results}
\begin{figure}[t]
\begin{center}
\includegraphics[width=\columnwidth]{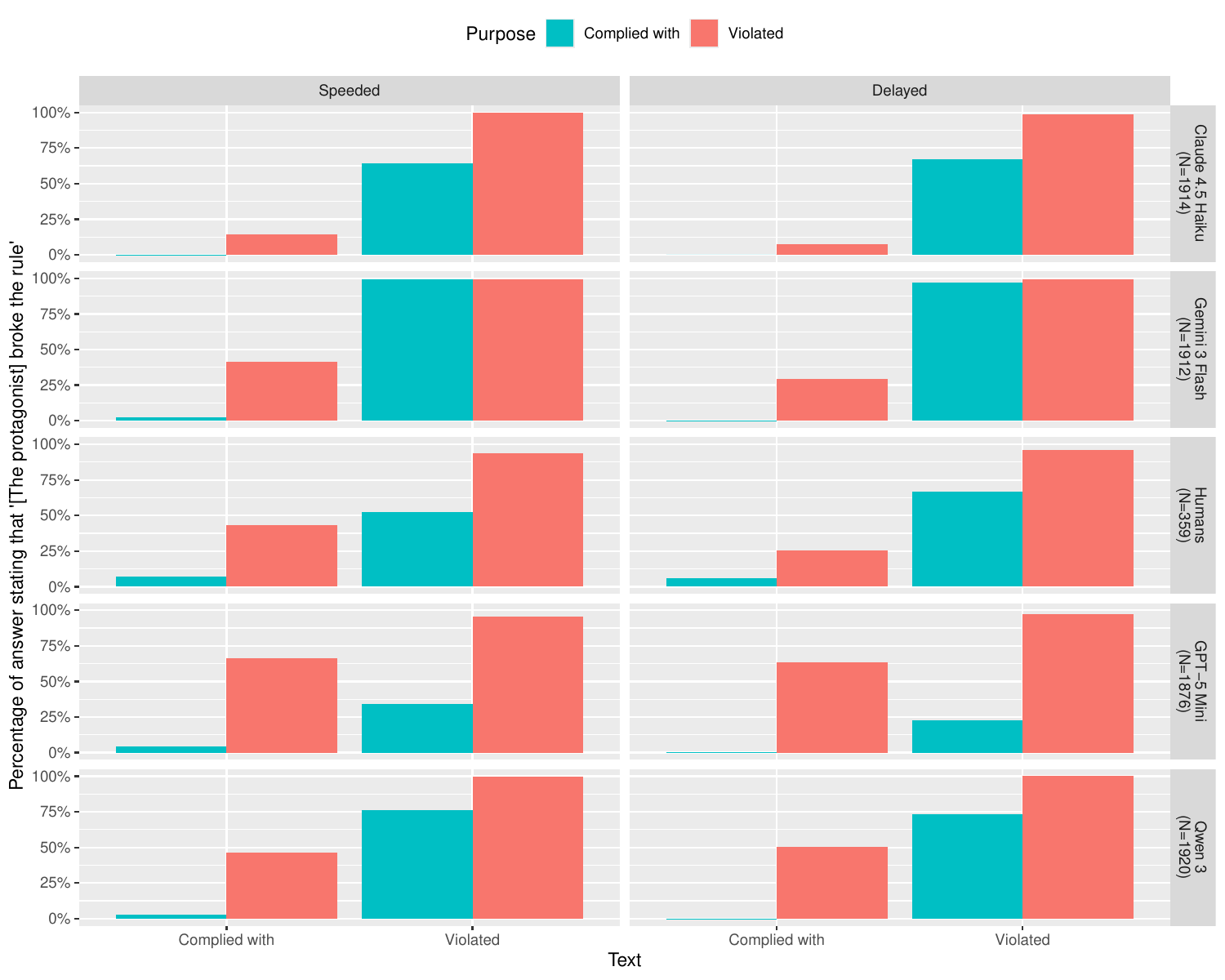}
\end{center}
\caption{Percentage of answers that agreed that the participant broke the rule by condition across all system prompts and scales.}
\label{fig:main-results-2b}
\end{figure}

Employing an empirical strategy similar to Study~1B, we fitted ANOVAs of text, purpose, condition, and all two and three-way interactions between them while accounting for random effects of model instance. For all LLMs, the main effects of text and purpose were significant (all $p\text{s} < .001$). Moreover, all models were affected by the time pressure manipulation. For Claude 4.5 Haiku, we observed significant main effects of condition ($F_{(1, 1651)} = 5.18, p = .023$) qualified by significant interactions with text ($F_{(1, 8400)} = 14.41, p < .001$) and purpose ($F_{(1, 8400)} = 22.29, p < .001$). Moreover, the significant interactions were in the same direction as those observed in humans, with amplified effects for text and dampened effects for purpose in the delayed condition compared to the speeded condition. For Qwen 3, we did not observe a significant main effect of condition ($F_{(1, 1681)} = 0.38, p = .537$); however, we did observe a significant interaction between condition and purpose ($F_{(1, 8473)} = 14.21, p < .001$), reflecting a larger effect of purpose in the delayed condition, which runs counter to the pattern observed among humans. For Gemini 3 Flash, we observed significant main effects of condition ($F_{(1, 1649)} = 53.95, p < .001$) qualified by significant interactions between condition and text ($F_{(1, 8535)} = 36.92, p < .001$), condition and purpose ($F_{(1, 8535)} = 13.47, p < .001$) and between condition, text, and purpose ($F_{(1, 8535)} = 38.27, p < .001$). Moreover, as with Claude 4.5 Haiku, these interactions ran in the same direction as among humans. Finally, for GPT-5 Mini we observed a significant main effect of condition ($F_{(1, 1640)} = 38.62, p < .001$) qualified by significant interactions between condition and purpose ($F_{(1, 8287)} = 26.02, p < .001$), reflecting a larger effect of purpose in the delayed condition (which is not what was observed among humans) and between condition, text, and purpose ($F_{(1, 8288)} = 20.68, p < .001$). These patterns are displayed in \cref{fig:main-results-2b}.

To understand the effect of prompt variation, we fitted the aforementioned ANOVA six times per model using different subsets: we stratified the data by each of the four levels of system prompt while collapsing across both scale presentations (with one of them reverse-coded) and by each of the levels of scale presentation while collapsing across all levels of system prompt. In other words, for each ANOVA we filtered the data for the relevant value of the prompt variation (instruction or scale) while dropping the other variation from the model. For all fitted models, the simple effects of text and purpose were significant (all $p\text{s} < .001$). For Claude, the interaction between text and condition was significant in four out of six subsets, as was the interaction between purpose and condition, while the three-way interaction was only significant once. For GPT~5 Mini, the main effects of condition were significant on three out of the six subsets, while in five its interaction with text and three-way interactions including purpose were significant. For Gemini~3 Flash, in a single case there was a main effect of condition, with five subsets displaying a significant interaction with purpose and the three-way interaction including text. Finally, for Qwen~3, only a single subset revealed a significant effect of condition qualified by its interaction with purpose, while another showed the same interaction with purpose and the three way interaction between text, purpose, and condition.

To evaluate to what extent LLM-generated answers in Study 2B aligned with original human-generated responses, we computed the mean rule violation judgment for each combination of text, purpose, condition for each model six times, again by subsetting the data by each of the four levels of system prompt while collapsing across both scale presentations and by each scale presentation while collapsing across system prompts. Overall, models were highly correlated with human responses across different subsets ($ 0.56 < r < 0.937; median(r)=0.776$, all $p_s < 0.002$).

\subsection{Discussion}

Study 2A showed that models diverge on how they react to manipulations that have a straightforward mechanistic route to influence rule violation judgments in humans, but not in LLMs: some models mimic the expected human behaviour, while others ignore the blocked manipulation. Study 2B applied the same setup of Study 2A to a more recent cohort of smaller LLMs while also testing for robustness through system prompt and scale endpoint variations.

Surprisingly, the insensitivity to the irrelevant manipulation observed in GPT-4o and Llama 3.2 90b was not replicated in this second cohort of models. In two cases (Claude 4.5 Haiku and Gemini 3 Flash), LLMs responded to the blocked manipulation in exactly the same way that humans did, raising concerns of memorisation. However, the other two LLMs (GPT-5-Mini, Qwen 3) responded to the time pressure manipulation in ways that differ from human reactions to the same stimuli. Finally, across the board, sensitivity to time pressure was substantially less robust to changes in system prompt and scale endpoint variation than the effects of text and purpose. Thus, while the smaller models tested in Study 2B seem to be more susceptible to the memorisation critique, the patterns which suggest this conclusion are substantially more brittle than the other patterns highlighted throughout this paper. Future research should compare across different sizes within model families to confirm whether conceptual mastery increases and memorisation decreases with parameter count.

\section{Study 3}

Among humans, a forced delay leads to more textualist judgments. This, however, doesn't necessarily mean that humans are sequentially computing a number of inferential steps to reach a nuanced conclusion. Instead, previous research has suggested that forced delay changes human responses simply by giving participants the opportunity to suppress an immediate response driven by morality \cite{FlanaganEtAl2023MoralAppraisalsGuide}. If that is indeed the mechanism through which a forced delay exerts influence, we shouldn't expect that giving humans even more time to reflect on their judgment would change things further. On the other hand, the only way to manipulate the amount of time an LLM has to think is by changing their ``inference time compute", a feature of so-called "reasoning'' models \cite{ShojaeeEtAl2025IllusionThinkingUnderstanding, XuEtAl2025LargeReasoningModels}. By examining whether LLM answers are robust to the amount of reasoning effort available to the model, we can test whether LLM conceptual competence resembles human fluency.

Reasoning models leverage chain-of-thought prompting to produce a variable number of intermediate reasoning tokens or hidden deliberative steps before generating their response. Interfaces for these models often include parameters that allows the user to adjust the maximum number of tokens available for reasoning, or settings that tune the model’s “thinking effort”. Effort will be controllable by levels such as low, medium and high -- sometimes with other options. Thus, while the mechanistic path through which time pressure and forced delay operate is blocked in the earlier cue-only studies, reasoning effort can be manipulated directly in reasoning models. Study 3 addresses whether such models adjust their rule violation judgments in response to variations in their thinking budget. If most models display little or no difference between low and high reasoning effort, this suggests that, as one would expect if the concept were fluently mastered, LLMs can apply rules by relying largely on their standing semantic competence rather than on expanded deliberative problem-solving.

We surveyed state-of-the-art models released by early July 2026 that explicitly supported adjustment of reasoning effort as a parameter. For this experiment, we adapted the stimuli and materials from \citet{FlanaganEtAl2023MoralAppraisalsGuide}, omitting instructions about the available time to answer, so that task context was held constant while the reasoning effort available to models was varied.

\subsection{Methods}

We produced 60 responses per reasoning effor level with each of the models included in this experiment, while randomly allocating one of the 12 groups that included 8 scenarios from Studies~2A and 2B. Our cohort of models for this experiment consisted of: GPT-5.4 Mini (\texttt{gpt-5.4-mini}), GPT-5.5 (\texttt{gpt-5.5}), Gemini 3.5 Flash ((\texttt{gemini-3.5-flash}), Claude Sonnet~5 (\texttt{claude-sonnet-5}), and GPT-OSS 120B (\texttt{gpt-oss-120b}), an open weight model from a commercial model provider which supports the effort level parameter. We requested each model through the provider Python API, except for GPT-Oss for which we used OpenRouter. We adapted the stimuli by removing the original instructions regarding the time available to answer, as well as the reminder we introduced after scenarios.

\subsubsection{Temperature Calibration}

As reasoning models need to produce traces before responding to prompts, developers often recommend not changing the temperature parameter, or simply do not make it available. As such, we opted to keep the temperature at the default level for each model, while varying the effort level between low, medium, and high.

\subsubsection{Data Processing}

We take the same measures to process data for this experiment, but discard the whole set of answers from the identified generation when one of the answers contained ambiguous or invalid responses. We found no answers failed the attention check, but discarded a total of 24 generations: 1 from Sonnet 5 on high effort, 2 from Gemini 3.5 on medium effort and 3 on high, and a total of 20 from GPT-OSS (10 on high, 5 on low, and 5 on medium effort).

\subsection{Results}
Following a procedure analogous to previous studies, we fitted individual ANOVAs with text, purpose, effort level and their two and three-way interactions, as well as random intercept for each model instance.
Throughout this analysis we treat effort level as an ordered factor ($low < medium < high$).

\begin{figure}[t]
\begin{center}

\includegraphics[width=\columnwidth]{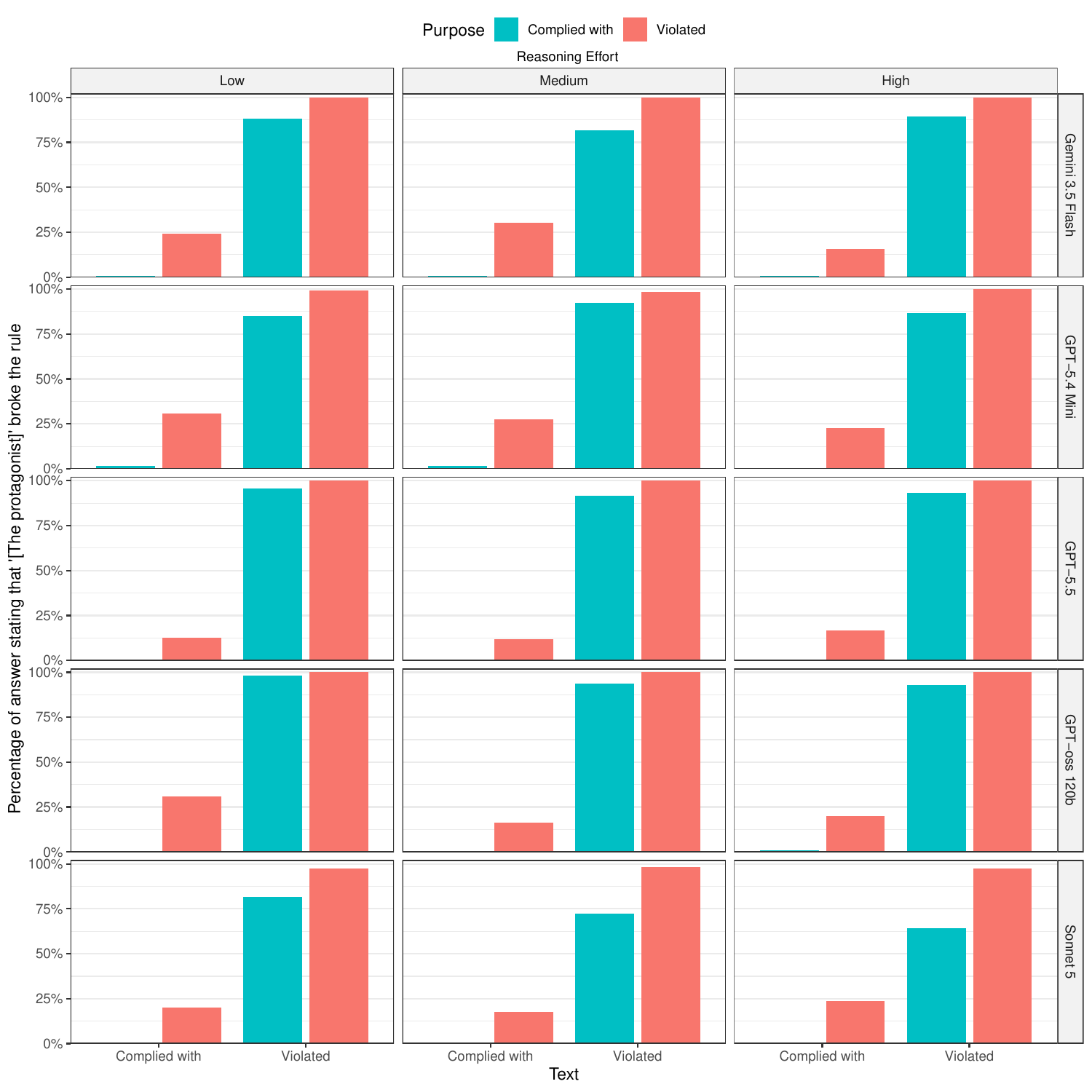}
\end{center}
\caption{Percentage of answers that agreed that the participant broke the rule by effort condition for each LLM.}
\label{fig:study3-effort}
\end{figure}

Only GPT-OSS showed a significant main effect of effort level ($p<0.005$) qualified by its three way interaction with text and purpose ($p<0.001$). To unpack these effects, we calculated Tukey adjusted pairwise contrasts, and found low effort present increased rule violation judgement compared to high ($b=0.38, p < 0.04$) and medium ($b=0.48, p < 0.005$) effort settings. The three way interaction indicates that the effect shifted between combinations of violation of text and purpose, as shown in \cref{fig:study3-effort}. 

Meanwhile, Sonnet displayed a significant interaction of effort level $\times$ text ($p = 0.027$). To decompose this effect, we tested whether the linear and quadratic coefficients of effort varied across conditions, which revealed a significant linear component ($p=0.009$). The effect of text violation decreased as thinking effort increased ($b_{low} = 0.792 > b_{medium} = 0.764 > b_{high} = 0.691$).

Finally, all results indicated stable effects of both text and purpose violation ($p_s < 0.001$) in their expected directions.
Their interaction was also significant for all models except Claude Sonnet ($p=0.12$).
% However, we found negative coefficients of the $text \times purpose$ interaction for Gemini 3.5 Flash and GPT 5.4 Mini.

We also fitted an overall ANOVA model that also included fixed effects for each model. This analysis revealed significant main effects of model, text, purpose ($p_s < 0.001$), and effort level ($p_s < 0.02$).
Marginal means for the trend analysis indicated increased effort had a significant negative effect on violation judgements ($b=-0.02, p=0.006$) while the quadratic component was not significant ($p=0.397$).
Importantly, we still see stable effects of text, purpose and their interaction in this experiment.
The effects of text and purpose were qualified by their two-way interaction and by significant two and three-interactions with model ($\text{all } p_s < 0.001$), while the four way interaction was trending to significance ($F=1.9$, $p=0.058$).
Effort level effect was qualified by a three-way $model \times effort \times text$ interaction, which likely tracks the individual model results reported, which found that effort only had significant main effect for GPT-Oss, and in its interaction with text for Sonnet~5.

% All interactions among model, text, and purpose were significant ($p_s < 0.001$), while the four way interaction was trending to significance ($F=1.9$, $p=0.058$). Effort level did not have a stable effect beyond its main effect, as none of the two way interaction were significant. It was, however, qualified by the model $\times$ effort $\times$ text interaction. These results are likely an artifact of the fact that effort only had significant main effect for GPT-Oss, and the interaction with text for Sonnet~5.

% To unpack the effect of thinking effort we estimated the linear and quadratic trends 
% This revealed significant negative linear trend ($b=-0.02$, $p=0.006$).
% Pairwise Tukey adjusted comparisons revealed a significant contrast between low and medium ($b=0.2$, $p=0.1$).
% Breaking down the effect of effort for model confirmed the negative linear trend was significative for GPT-oss ($b=-0.038$, $p=0.35$), and Sonnet ($b=-0.034$, $p=0.43$).
% Follow up pairwise Tukey adjusted comparisons for these models, indicated only the medium-low contrast for GPT-oss was significant ($b=0.048$, $p=0.017$

\subsection{Discussion}

There is reason to think that, when mastered, the concept of rule is one that should be applied fluently rather than through the explicit concatenation of several inferential steps. While the overall model indicated a significant effect of effort level, for most models, additional deliberative effort did not significantly alter rule-application judgments, supporting the hypothesis that the concept was already available for fluent deployment through standing semantic competence
\footnote{We ran a qualitative exploration by running 3 extra runs for each model-\texttt{effort-level} pair, while saving and including reasoning-traces in context, and sampling them from GPT-Oss since they were retained for all Study~3 responses. This is available in our OSF Supplementary Materials. Although sample size was low, we found that adaptative thinking made models unreliable in generating output traces, even if effort level increased the amount of responses with reasoning. Gemini~3.5 Flash produced them for most queries, whereas Claude Sonnet produced only 3 in respect of its 24 responses under high effort. Qualitative coding of the reasoning traces indicated a tendency to display textualist reasoning even for cases where only purpose had been violated, with few traces acknowledging rules' purpose.}.
The modest but significant effects of reasoning effort for GPT-OSS and Sonnet show, however, that this pattern of non-deliberative concept application is not uniform across reasoning models.   
  
%%%%%%%%%%%%%%%%%%%%%% DISCUSSION %%%%%%%%%%%%%%%%%%%%%%
\section{General Discussion}
Are the impressive results achieved by LLMs on cognitive science experiments caused by memorization? Our experiments suggest that this is not the case. LLM responses track human judgment in novel stimuli that could not have been part of the pre-training data (Study~1A), a result that was robust even if the data may have been partially available for training (Study~1b).

In Studies~1A and 1B, both humans and LLMs became slightly, but significantly, less wedded to rule text on the newly crafted vignettes. This, we argue, is also evidence of conceptual competence. We designed the vignettes to be perfectly matched and confidently pre-registered the prediction that levels of adherence to text and purpose would correspond across old and new vignettes. However, the data proved us wrong. Both humans and LLMs tapped into some subtle systematic difference in the stimuli that led to decreased textualism. That LLMs reacted to this subtle difference just as humans did is surprising and points to an implicit facility with the concept. 

An alternative concern focuses on brittleness. If the responses of LLMs are sensitive to extraneous variation in the employed system prompt, or of the meaning attached to numeric values in a psychological scale, then it seems hard to accept that they have concept mastery. Studies 1B and 2B found that they are not: the drivers of rule violation judgment remain largely the same across different system prompts and numerical anchors.

A different way to test for memorization leverages the fact that some pathways to influence rule violation judgments are available in human experiments, but unavailable for LLMs. In Studies 2A and 2B we focused on one such pathway: time pressure. When time is limited, humans are more likely to prioritise a rule's purpose over its letter than when given more time to think. However, when forced to delay their responses for a few seconds, humans suppress this moralistic inclination, leading to more textualist responses (which might be plausibly attributable to Resource Rational processes \cite{IcardmanuscriptResourceRationality, LiederEtAl2026RationalUseCognitive}). Meanwhile, LLMs are not subject to the same constraints at run-time in our studies. Thus, if conceptual competence requires sensitivity to relevant differences and robustness to irrelevant ones, competent models should ignore the mentions to time in the prompts.

The results of Studies 2a and 2b were mixed in this respect. In some cases the extraneous time pressure was significant, while in others it was not. Notably, replication of the pattern observed in humans was most apparent in models with fewer parameters (Study 2B). We interpret these results, together with those of human participants, as indicating a distinction between LLM competence in a concept's application and strict alignment with its application by humans. Models that track the case-internal features of rule application while ignoring a merely described time constraint may be displaying the kind of robustness one would expect from human concept competence, even though their responses are less human-like. 

Evidence of LLM conceptual competence extended to variation in reasoning settings. Most models did not significantly vary their rule applications across different levels of deliberative capacity, while retaining sensitivity to both the spirit and letter of rules. This pattern mirrors evidence that human conceptual competence in this domain is fluent rather than deliberatively reconstructed. For both humans and LLMs, rule application appears to draw primarily on standing semantic competence rather than expanded deliberative processing.

Finally, our efforts to match variance in responses from LLMs to human samples proved insufficient. Despite choosing a temperature that most approximated variance in the set of old vignettes, when prompted on a larger batch, including the new scenarios, LLMs still displayed significantly less variation.

\subsection{Implications}

As the application of chatbots in courts proliferates \cite{gutierrez_unesco_2024}, some judges are publicly calling attention to AI's potential to assist legal reasoning:``[Judges] should consider whether and how AI-powered large language models\dots\ might\dots\ inform the interpretive analysis'' \cite{newsom_snell_2024}. The growth in judicial interest in LLMs has in turn prompted worries that judges' reliance on this technology may ``impact normative ideals of how justice should be done'' \cite[656]{barry_how_2023}. Traditionally, the main disadvantage of using AI systems in applying rules has been thought to be their inability to identify novel, yet intuitively relevant case features: ``[h]uman judges and other persons charged with interpreting legal texts reason in ways that\dots\ remain over the horizon of machine capacities'' \cite[239]{whalen_rule_2020}; similarly \cite[260]{re_developing_2019}. Our research indicates that this may no longer be the case. On the evidence of the reported studies, the absence of a human touch need not necessarily degrade the quality of rule violation judgements. 

More generally, across five experiments, we have shown that rule violation judgments in at least some LLMs respond to the same general cues as those rendered by humans and preserve the central effects of rule text and purpose across novel vignettes, variations in system-prompt wording and numerical anchors, and changes in irrelevant instructions. This evidence tends to undercut simple memorisation accounts. At least with regards to LLMs' application of rules, some genuine conceptual competence seems to be at play. Future research should take this as a starting point to move the debate forward.

For instance: LLMs' concrete-level judgments are sometimes uncorrelated with their abstract-level traits \cite{jung-etal-2026-psychometric, nunes_are_2024}. If LLMs were sophisticated memorisation machines, this could perhaps be easily explained: for some reason, it could be that abstract-level training materials were leaning towards profile A, while most concrete-level materials recommended responses incompatible with profile A. Given, however, that LLMs are doing at least some generalisation, we need an alternative explanation for this and other systematic differences between humans and LLMs.

\section{Acknowledgments}

AI tools were used to assist in writing the code for this paper and generating its figures. Human Data Collection carried for this paper was authorized by ERB protocol XXXX..

\printbibliography

\appendix

\section{Summary of models used in the paper}

% \usepackage{multirow}
\begin{table}[h]
\caption{Overview of models used in our paper, their specific version and data training cutoff}
\label{tab:models-overview}
\begin{tabular}{llll}
\hline
\textbf{Study}          & \textbf{Model}       & \textbf{Version}                                & \textbf{Training Data Cutoff} \\ \hline
\multirow{4}{*}{1a, 2a} & GPT-4o                & \texttt{gpt-4o-2024-08-06}     & October, 2023                 \\
                        & LLAMA~3              & \texttt{Llama 3.2 90B Instruct}    & December, 2023              \\
                        & Claude~3             & \texttt{Claude 3 Opus 20240229}    & August, 2023                  \\
                        & Gemini~Pro               & \texttt{Gemini 1.5 Pro 002}        & May, 2024                     \\ \hline
\multirow{4}{*}{1b, 2b} & Qwen~3          & \texttt{qwen3-235b-a22b-2507}      & June, 2025 (Source from openrouter not developer)                    \\
                        & Claude 4.5 Haiku     & \texttt{claude-haiku-4-5-20251001} & July, 2025                \\
                        & GPT-5 Mini           & \texttt{gpt-5-mini-2025-08-07}     & May, 2024                     \\
                        & Gemini~3 Flash  & \texttt{gemini-3-flash-preview}    & January, 2025                 \\ \hline
\multirow{5}{*}{3}      & GPT-5.4 Mini         & \texttt{gpt-5.4-mini}              & August, 2025                  \\
                        & GPT-5.5              & \texttt{gpt-5.5}                   & December, 2025                    \\
                        & Gemini 3.5 Flash     & \texttt{gemini-3.5-flash}          & January, 2025                 \\
                        & Claude Sonnet~5 & \texttt{claude-sonnet-5}           & January, 2026                 \\
                        & GPT-OSS 120B         & \texttt{gpt-oss-120b}              & June, 2024                    \\ \hline
\end{tabular}
\end{table}

\end{document}